\pdfoutput=1

\documentclass[11pt]{article}

\usepackage[]{emnlp2021}
\usepackage{times}
\usepackage{latexsym}

\usepackage[T1]{fontenc}

\usepackage[utf8]{inputenc}

\usepackage{graphicx}
\usepackage{subfigure}
\usepackage{amssymb,amsmath,latexsym,amsfonts,amsthm,cleveref, bm, bbm}
\usepackage{stmaryrd}
\usepackage{multirow}
\usepackage{stfloats}
\usepackage{booktabs}
\usepackage{color}

\usepackage{microtype}

\title{Towards Document-Level Paraphrase Generation \\with Sentence Rewriting and Reordering }

\author{First Author \\
  Affiliation / Address line 1 \\
  Affiliation / Address line 2 \\
  Affiliation / Address line 3 \\
  \texttt{email@domain} \\\And
  Second Author \\
  Affiliation / Address line 1 \\
  Affiliation / Address line 2 \\
  Affiliation / Address line 3 \\
  \texttt{email@domain} \\}
\author{Zhe Lin, Yitao Cai \and Xiaojun Wan \\
 Wangxuan Institute of Computer Technology, Peking University \\
 Center for Data Science, Peking University \\
 The MOE Key Laboratory of Computational Linguistics, Peking University \\
 {\tt \{linzhe,caiyitao,wanxiaojun\}@pku.edu.cn} \\}

\begin{document}
\maketitle
\begin{abstract}

Paraphrase generation is an important task in natural language processing. Previous works focus on sentence-level paraphrase generation, while ignoring document-level paraphrase generation, which is a more challenging and valuable task. In this paper, we explore the task of document-level paraphrase generation for the first time and focus on the inter-sentence diversity by considering sentence rewriting and reordering. We propose \textbf{CoRPG} (\textbf{Co}herence \textbf{R}elationship guided \textbf{P}araphrase \textbf{G}eneration), which leverages graph GRU to encode the coherence relationship graph and get the coherence-aware representation for each sentence, which can be used for re-arranging the multiple (possibly modified) input sentences. We create a pseudo document-level paraphrase dataset for training CoRPG. Automatic evaluation results show CoRPG outperforms several strong baseline models on the BERTScore and diversity scores. Human evaluation also shows our model can generate document paraphrase with more diversity and semantic preservation.
\end{abstract}

\section{Introduction}
Paraphrase generation \citep{mckeown-1983-paraphrasing, 10.3115/1073445.1073448} is an important task in natural language processing, and it aims to rewrite a text in other forms while preserving original semantics. Paraphrase generation has many applications in other down-stream tasks, such as text summarization \citep{JointCopy}, dialogue system, question answering \citep{xu-etal-2016-paraphrase}, semantic parsing \citep{berant-liang-2014-semantic} and so on.
Inspired by the success of deep learning, 
most paraphrase systems leverage existing paraphrase corpora to train a seq2seq model, such as variational auto-encoder \citep{gupta2017deep}, syntactic pre-ordering \citep{goyal-durrett-2020-neural} and so on. 
All these works focus on sentence-level paraphrase generation. 

\begin{table}
    \centering
    \small
    \begin{tabular}{|p{7cm}|}
    \hline
    \textbf{Original:} \cr
    \textcolor{red}{$_1$}Sustainability has become the foundation for almost all economic thinking nowadays. 
    \textcolor{blue}{$_2$}It is essential not only to economic recovery today, but to ensuring peace and security tomorrow.
    \textcolor{olive}{$_3$}Factoring sustainability into all our thinking is necessary because, as a global society, we are living on the edge.
    \textcolor{orange}{$_4$}The last two years have brought a series of crises: energy, food, climate change, and global recession.
    \textcolor{purple}{$_5$}I fear that worse may be in store.\cr
    \hline
    \textbf{Paraphrase:} \cr
    \textcolor{red}{$_1$}Today, sustainability has been the basement for almost all economic mind.
    \textcolor{olive}{$_3$}It is necessary to reflect on sustainable development in our planning, \textcolor{blue}{$_2$}because it is indispensability not only for current economic recovery, but for the peace and security tomorrow.
    \textcolor{olive}{$_3$}$_,$\textcolor{orange}{$_4$}The global society has experienced a series of crises in the past two years, such as energy, food, climate change and global economic recession. 
    \textcolor{olive}{$_3$}We have on the edge of collapse, \textcolor{purple}{$_5$}but I worry that the worse things are yet to come. \cr
    \hline
    \end{tabular}
\caption{\small An example for sentence reordering, splitting and merging in document paraphrase. The number before each sentence in the paraphrased document indicates the corresponding original sentence from the input document. }
\label{example}
\end{table}

Document-level paraphrase generation, which aims to rewrite a passage or a document without changing its original meaning, is a more valuable and challenging task. However, because of the lack of parallel corpora, there is few research on document-level paraphrase generation.
The difference between sentence-level paraphrase generation and document-level paraphrase generation is that the former task only focuses on the lexical and syntactic diversity of a sentence, while the latter task also needs to introduce the diversity across multiple sentences (we call it \textbf{inter-sentence diversity}), such as sentence reordering, sentence merging and splitting. 
Sentence reordering is to reorder the sentences without significantly deteriorating the coherence of the document. Sentence merging and splitting aim to merge two or more sentences into one sentence, and vice versa. An example about document-level paraphrase is shown in Table \ref{example}. As is shown in this example, there is inter-sentence diversity in paraphrase. For example, the third sentence in original document can be decomposed and correspond to three parts in the paraphrased document (i.e., the main clause of the second sentence, the first words of the third sentence and the first words of the last sentence). Each of the last three sentences in the paraphrase is composed by merging multiple sentences in the original document, and the way of narration has been changed. These operations can effectively improve the diversity of document paraphrase, but they are beyond the ability of sentence-level paraphrasing model.

In this work, \textbf{we conduct a pilot study of this challenging task and focus only on rewriting and reordering the sentences in original document} while still maintaining the original semantics and inter-sentence coherence. Due to the lack of parallel document-level paraphrase pairs, it is not possible to straightforwardly train a sequence-to-sequence paraphrasing model to address this task. We thus propose \textbf{CoRPG} (\textbf{Co}herence \textbf{R}elationship guided \textbf{P}araphrase \textbf{G}eneration), which is based on an automatically constructed pseudo document paraphrase dataset. Though the paraphrases in the pseudo dataset do not involve inter-sentence diversity, our model can learn the coherence relations between sentences via a coherence relationship graph generated by ALBERT \citep{Lan2020ALBERT:}, and make use of the learned coherence-aware representations of sentences to reorder them, while keeping good coherence of the generated document. 


Our model consists of three parts: sentence encoder, graph GRU and decoder. Sentence encoder only encodes each sentence in the document individually. We propose graph GRU, which combines graph attention \citep{velickovic2018graph} and GRU, to catch the coherence relationship information. Finally, the outputs of graph GRU and sentence encoder are concatenated and used as input to decoder to generate the paraphrase. Extensive evaluations are performed and our model gets the best scores on most metrics in both automatic evaluation and human evaluation.

The contributions of our work are summarized as below:  

1) To the best of our knowledge, we are the first to explore the problem of document-level paraphrase generation and point out the difference between document-level paraphrase and sentence-level paraphrase.

2)We propose a new model \textbf{CoRPG} to address both sentence rewriting and reordering for document-level paraphrase generation. Our model can leverage graph GRU to learn coherence-aware representations of sentences and re-arrange the input sentences to improve the inter-sentence diversity of generated document paraphrases.

3) Both automatic evaluation and human evaluation show that our model can generate document paraphrase with high diversity, semantic relevance and coherence. Our code is publicly available at \url{https://github.com/L-Zhe/CoRPG}.

\section{Related Work}
With the development of deep learning, most paraphrasing models are based on seq2seq model.  
\citet{prakash2016neural} leveraged stacked residual LSTM networks to generate paraphrases. \citet{gupta2017deep} found deep generative model such as variational auto-encoder can improve the quality of paraphrase significantly. \citet{li-etal-2019-decomposable} proposed DNPG to decompose a sentence into sentence-level pattern and phrase-level pattern to make neural paraphrase generation more controllable. 
\citet{goyal-durrett-2020-neural} used syntactic transformations to softly “reorder” the source sentence and guide neural model to generate more diverse paraphrase. \citet{kazemnejad-etal-2020-paraphrase} explored to generate paraphrase by editing the original sentence.

Beside, there is another way to generate paraphrase, called ``pivoting", which leverages back-translation to introduce diversity. 
Recently, \citet{mallinson-etal-2017-paraphrasing} revisited this method with neural machine translation
to improve the paraphrase quality. \citet{wieting-gimpel-2018-paranmt} leveraged bidirectional translation model to construct paraNMT, which is a very large sentence-level paraphrase dataset.

All works above focus on sentence-level paraphrase generation, and to the best of our knowledge, there is no research on document-level paraphrase generation.

\section{Our CoRPG Model}

\subsection{Overview and Notations}
\begin{figure*}[htb]
\centering
\includegraphics[scale=0.57]{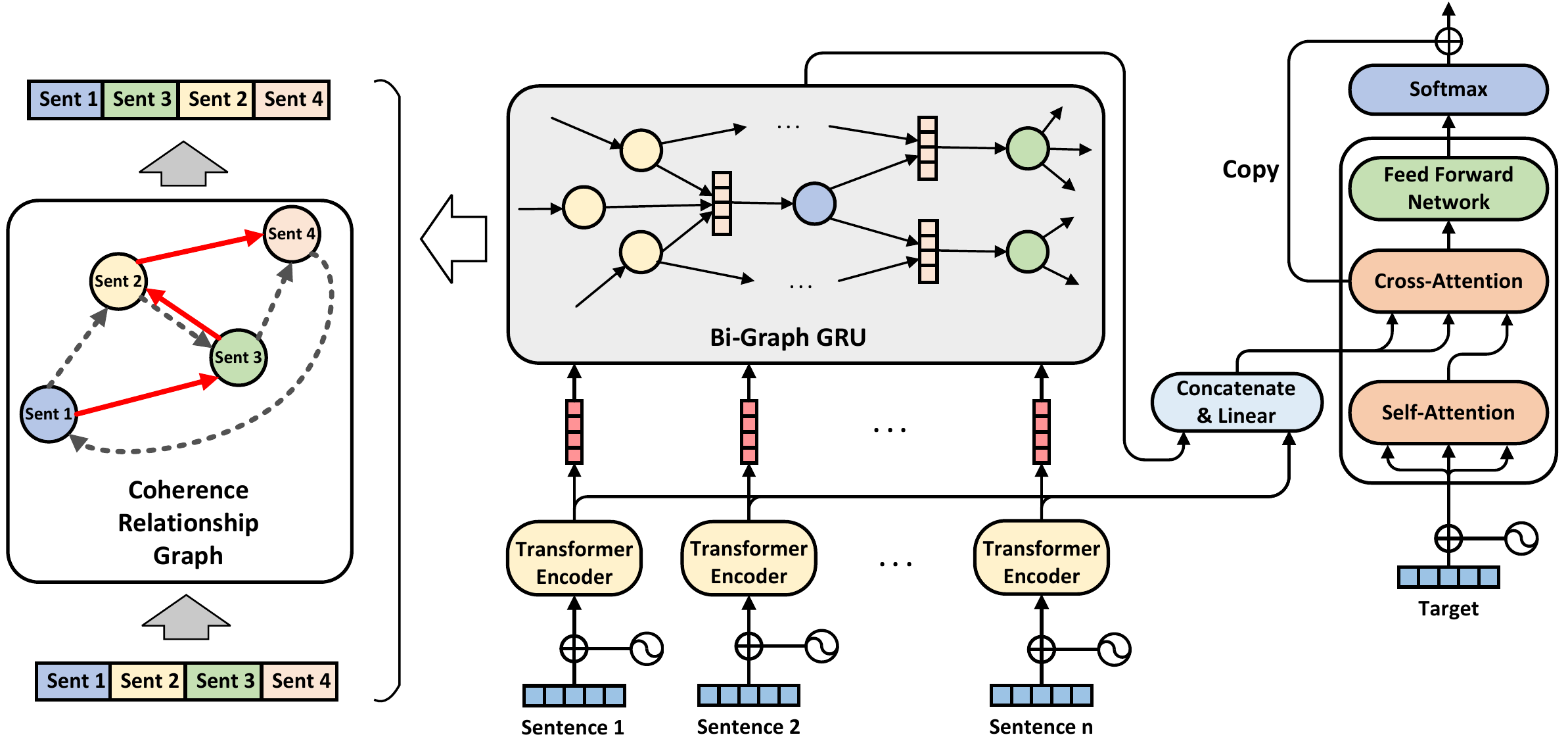}
\caption{ An overview of CoRPG, which consists of sentence encoder, graph GRU and decoder. }
\label{overview}
\end{figure*}
\subsubsection{Model Overview}

Figure \ref{overview} shows the overview of our model, which consists of a sentence encoder, a graph GRU and a decoder. Given an input document, we use the sentence encoder to get the representation of each sentence in the document, while ignoring the positional information of the sentence. We construct a coherence relationship graph for the sentences and use the graph GRU to get the coherence-aware representation of each sentence. The outputs of the sentence encoder and the graph GRU are taken by the decoder for generating a coherent paraphrased document. Note that we do not have a document paraphrase dataset with inter-sentence diversity. Instead, we use a sentence paraphrasing model to construct a pseudo document paraphrase dataset with only intra-sentence diversity (i.e., lexical or syntactic diversity within each sentence).  Our model is trained on this dataset to have the ability to reconstruct a coherent document by modifying and arranging multiple input sentences without using the original positional information of the sentences\footnote{If we use the original positional information of input sentences for training, the model can simply output a document with the same positions of these sentences. During testing, the model cannot generate document paraphrase with sentence reordering and rearrangement.}. In other words, the input to our model can be seen as just a set of sentences without sequential order. The key to achieving this is the coherence-aware representations of the sentences learned by the graph GRU. During testing, our model can re-organize the text according to the learned coherence-aware and semantic representations of the sentences. The output document is very likely to have different sentence ordering and arrangement, as compared to the input document, because there are usually different reasonable ways for arranging a set of sentences, besides the original sentence order. The details of the dataset and model modules will be given in the next sections.

\subsubsection{Pseudo Document-Level Paraphrase Dataset}
Our model regards paraphrase as a monolingual translation task. Given a document $D = \{S^1, S^2, \cdots, S^N\}$, where $N$ is the total number of sentences in the document and $S^i$ is the $i$-th sentence in the document. Because of lack of gold document paraphrase dataset, we leverage an off-the-shelf sentence paraphrasing model to generate a pseudo document paraphrase $D_p = \{S^1_p, S^2_p \cdots, S^N_p\}$ sentence by sentence, where $S^i_p$ is obtained by paraphrasing $S^i$ with the sentence paraphrasing model. Then, we set $D_p$ as input and $D$ as target to train our model.

\subsubsection{Coherence Relationship Graph}
There are many works focusing on text coherence, such as NCOH \citep{moon-etal-2019-unified} and CohEval \citep{Mohiuddin2020CohEvalBC}. Many pre-trained models also introduce text coherence as a subtask to improve the generalization ability. For example, \citet{devlin-etal-2019-bert} employed next sentence prediction (NSP) task to train BERT; \citet{Lan2020ALBERT:} proposed ALBERT, which leverages sentence order prediction (SOP) to catch the inter-sentence coherence better. We employ the SOP probability of ALBERT to measure the inter-sentence coherence. The coherence relationship graph $\mathbb{G}$ for document $D_p = \{S^1_p, S^2_p, \cdots, S^N_p\}$ takes sentences as nodes and the edge is defined as follows:

\begin{equation}
    \mathbb{G}(i, j) = \left\{
\begin{aligned}
        \mathbbmss{1}\{P_{SOP}&(S_p^i, S_p^j) \ge \epsilon \} \footnotemark[1] \, & i \neq j \\
    &0 & i = j
\end{aligned}
\right.
\end{equation}
\footnotetext[1]{$\mathbbmss{1}\{\cdot\} = 1$ if $\cdot$ is true. Otherwise it equals to $0$.}

\noindent where $P_{SOP}$ is the SOP probability of ALBERT. $\mathbb{G}(i, j) = 1$ means that it is coherent to put the $i$-th sentence before the $j$-th sentence.


\subsection{Sentence Encoder}
Sentence encoder aims to obtain each sentence's contextual representations in a document. We choose the Transformer encoder \citep{vaswani2017attention} as our sentence encoder. 

Each sentence in the document is sent to the sentence encoder respectively. For sentence $S^i$ (Actually $S^i_p$, but for simplicity we omit the subscript $p$ here), we obtain the output encoding matrix as $S^i_e = \{h^i_1, h^i_2, \cdots, h^i_n \}$, where $h^i_j \in \mathbb{R}^{d_{model}}$ is the word embedding vector and $n$ is the number of words in $S^i$. Then, the outputs of sentence encoder for the whole document is $D_e = \{S^1_e, S^2_e, \cdots, S^N_e\}$. Notice that, because we encode each sentence individually and we do not encode the positional information of each sentence, we assume the learned sentence representations only contain semantic information, but no (or very little) coherence or sequential information.

Same as \citep{cao-etal-2020-jointly}, we get the sentence vector by averaging the word embedding vectors of this sentence. Then, the sentence representation matrix of the document is $D_r = \{r^1, r^2, \cdots, r^N \}$, where $r^i = \frac{1}{n}\sum_{j=1}^n h^i_j$.

\subsection{Graph GRU}
\begin{figure*}[htb]
\centering
\includegraphics[scale=0.5]{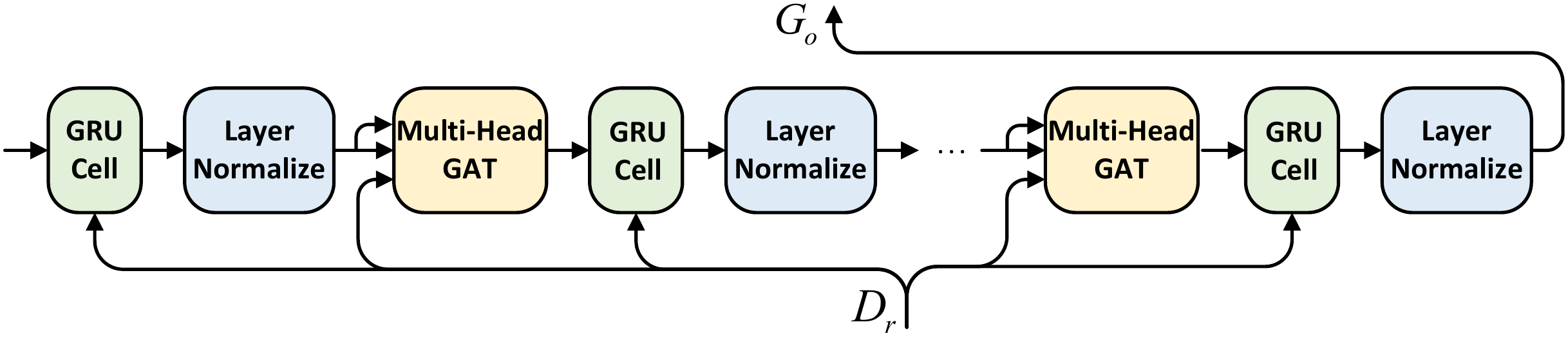}
\caption{\small The structure of graph GRU, which is a stack of $L_g$ identical layers. Each layer includes a multi-head graph attention block, a GRU cell and a layer normalization. }
\label{graph_gru}
\end{figure*}

Most of the previous graph models focus on encoding the semantic information in the graph \citep{beck-etal-2018-graph,guo-etal-2019-densely} or leveraging graph information to guide sequence encoding or decoding \citep{peng-etal-2017-cross}. However, few work focuses on the coherence relationship in the graph. In this section, we propose the graph GRU to explore the coherence relationship between sentences. We assume there exists coherence relationship between $S^i$ and $S^j$ if $\mathbb{G}(i, j) = 1$. For a sentence in the coherence relationship graph, there may be more than one precursor, and we leverage graph attention to aggregate the information from all its precursors. Then, we regard this information as hidden information in GRU cell \citep{cho-etal-2014-learning} to catch the coherence relationship. Figure \ref{graph_gru} shows the structure of the graph GRU.

Our graph GRU is a stack of $L_g$ identical layers. Each layer includes a multi-head graph attention block, a GRU cell and a layer normalization. All layers share the same parameters. For normalizing the input of each layer, we leverage zero vector instead of the graph attention vector as the hidden information of the GRU cell in the first layer. The input to the graph GRU is $D_r$. We denote the output of $l$-th layer as $G_l = \{g^l_1, g^l_2, \cdots, g^l_N\}$, where $g^l_i \in \mathbb{R}^{d_{model}}$ is the representation of the $i$-th node in the graph. We will describe the graph GRU in detail.

First, we define a graph attention operation. Graph attention \citep{velickovic2018graph} is used to aggregate the information from neighbor nodes. We calculate the graph attention between sentence vectors $D_r$ and the outputs of $l$-th layer $G_l$. For simplicity and clarity, we omit the layer index $l$ for nodes. The aggregate operation is as follow:
 

\begin{equation}
    \operatorname{GAT} (r_i, G) = \sum_{\substack{\mathbb{G}(i, j) = 1 \\ g_j \in G}} \alpha_{ij} g_j \mathbf{W}_V
\end{equation}

\noindent where $\mathbf{W}_V \in \mathbb{R}^{d_{model} \times d_u }$. $\alpha_{ij}$ is the attention coefficient computed as follow:

\begin{equation}
\begin{aligned}
    s_{ij} &= \left(r_i \mathbf{W}_Q \right) \left(g_j \mathbf{W}_K\right)^{\top} \\
    \alpha_{ij} &= \frac{\operatorname{exp} \left( s_{ij} \right)}{\sum_{\mathbb{G}(i, k) = 1} \operatorname{exp} \left( s_{ik} \right)}
\end{aligned}
\end{equation}

\noindent where $\mathbf{W}_Q, \mathbf{W}_K \in \mathbb{R}^{d_{model} \times d_{u}}$ are learnable parameters. Notice that, there exists sink node in the coherence relationship graph. If the $i$-th node is sink node, then $\mathbb{G}(i, \cdot) = 0$. For all sink nodes, we set all their attention weight $\alpha_{i\cdot} = 0$.

For better performance, we introduce a multi-head operation in graph attention.

\begin{equation}
\begin{aligned}
    \hat{g}_i &= \operatorname{GAT}(r_i, G) \quad r_i \in D_r\\
    \operatorname{Head}_j &= (\hat{g}_1, \hat{g}_2, \cdots, \hat{g}_N) \\
    \operatorname{MHGAT}(D_r, G) &= \left(\bigparallel_{j=1}^H  \operatorname{Head}_j \right)\mathbf{W}_{o}
\end{aligned}
\end{equation}

\noindent where $H$ is the head number, $\bigparallel$ is the concatenate operation, $\mathbf{W}_o \in \mathbb{R}^{H*d_{u} \times d_{model}}$. 

$G_{l-1}$ contains the coherence information of nodes with length $l-1$. We employ multi-head graph attention to aggregate the precursor node information of each node, and send the aggregated vector into GRU cell as hidden information. The details are as follow:

\begin{equation}
\begin{aligned}
    \bar{G}_l &= \operatorname{MHGAT} \left(D_r, G_{l-1}\right) \\
    z_t &= \sigma \left(\left[\bar{G}_l\bigparallel D_r \right] \mathbf{W}_z \right) \\
    r_t &= \sigma \left(\left[\bar{G}_l\bigparallel D_r \right] \mathbf{W}_r \right) \\
    \tilde{G_l} &= \operatorname{tanh}\left(\left[r_t\otimes\bar{G}_l\bigparallel D_r \right] \mathbf{W}_m \right) \\
    \hat{G}_l &= \left(1 - z_t\right) \otimes \bar{G}_l + z_t \otimes \tilde{G}_l \\
\end{aligned}
\end{equation}

\noindent where $\sigma$ is the sigmoid activation function, $\otimes$ is the element-wise product between matrices, and $\mathbf{W}_z, \mathbf{W}_r, \mathbf{W}_m \in \mathbb{R}^{2d_{model} \times d_{model}}$.

Finally, we leverage layer normalization \citep{Ba2016LayerN} to normalize $\hat{G}_l$. The outputs of $l$-th graph GRU layer is as follow:

\begin{equation}
\begin{aligned}
    G_l = \operatorname{LayerNorm} (\hat{G}_l)
\end{aligned}
\end{equation}

Different from traditional RNN model which cycles through each token in the sequence, our graph GRU encodes the coherence information by multi-layer propagation. Each new layer will increase the length of the encoding sequence by one. Therefore, the number of graph GRU layer $L_g$ is not a fixed number, but equals to the number of sentences in the document. We take the output of the last layer as its final output.

Following the idea of bidirectional RNN, we adopt two graph GRUs which do not share parameters to aggregate the coherence information in both directions. We send $\mathbb{G}$ into forward graph GRU and $\mathbb{G}^{\top}$ into reversed graph GRU, and get their outputs $\mathop{G} \limits ^{\rightarrow}$ and $ \mathop{G} \limits ^{\leftarrow}$ respectively. Finally, we combine the outputs in two directions as the final output of our bi-graph GRU.

\begin{equation}
    G_o = \mathop{G} \limits ^{\rightarrow} +  \mathop{G} \limits ^{\leftarrow}
\end{equation}

\noindent where $G_o = \{g_1, g_2, \cdots, g_N\}$, $g_i \in \mathbb{R}^{d_{model}}$ is the sentence vector containing coherence relationship information.

\subsection{Decoder}

We leverage Transformer decoder as our decoder. First, we combine the outputs of sentence encoder and graph GRU.

\begin{equation}
\begin{aligned}
    S^i_c &= \left[h^i_j \bigparallel g_i \right]_{j = 1}^n \\
    \tilde{d}_{c} &= \left[S^1_c, S^2_c, \cdots, S^N_c\right] \mathbf{W}_c + \mathbf{b}_c \\
    d_c &= \operatorname{LayerNorm}\left(\operatorname{ReLU}\left( \tilde{d}_c \right)\right)
\end{aligned}
\end{equation}

\noindent where $\mathbf{W}_c \in \mathbb{R}^{2d_{model} \times d_{model}}$, $\mathbf{b}_c \in \mathbb{R}^{d_{model}}$, $g_i \in G_o$. In order to avoid overfitting, we add dropout after ReLU function.
The combination operation above can be regarded as introducing the inter-sentence coherence relationship information to each sentence embedding matrix. 

Then, we send $d_c$ into decoder to guide the generation. 
We add copy mechanism\citep{see-etal-2017-get}. We leverage the average attention weight over all heads in the last decoder layer as the copy probability to calculate the final output's probability.




\subsection{Diversity Coefficient}
\label{sec:div_coef}
During experiment, we find that paraphrase model tends to copy original sentence. Therefore, our pseudo document paraphrase dataset created by sentence-level paraphrasing model has less diversity on both lexical and syntactic than the original sentence paraphrase dataset. To tackle this problem, we introduce diversity coefficient to pay more attention on diversity of N-gram phrase. 

We define the set of all N-gram phrases of source document as $U_N$. For a word $w$ in target document, we define the set of all N-gram phrases containing this word as $W_N$. Then, the loss of $w$ is as follow:

\begin{equation}
\begin{aligned}
    \tilde{I}_N &= \mathbbmss{1} \left\{ U_N \cap W_N = \varnothing \right\} \\
    I_N &= \tilde{I}_N \wedge \mathbbmss{1} \left\{ \sum_{i<N} I_i = 0 \right\} \\
    loss_w &= -\operatorname{log} P(w) \times \left(1 + \sum_N I_N\lambda_N \right) 
    \label{div coef}
\end{aligned}
\end{equation}

\noindent where $P(w)$ is the generation probability of $w$, $\lambda_N$ is a hyper-parameter which measures how much attention should be paid to N-gram diversity. Appx.\ref{appendix::div-coef} shows the detailed explanation of Eq.\ref{div coef}.

\section{Experiments}
\subsection{Datasets}

Because there is no gold document-level paraphrase dataset, we leverage sentence-level paraphrase dataset to train a sentence-level paraphrasing model and use it to generate pseudo document-level paraphrase dataset by paraphrasing every sentence individually in given documents. For sentence-level paraphrase dataset, we leverage \textbf{paraNMT} \citep{wieting-gimpel-2018-paranmt} \footnote{\url{https://www.cs.cmu.edu/˜jwieting}}. For document dataset, we employ \textbf{News Commentary} \footnote{\url{http://www.statmt.org/wmt20/translation-task.html}} which has been used in document-level machine translation. We sample 3000 documents (without references) from News Commentary for test.
Appx.\ref{appendix:dataset} shows more details about the data.

\subsection{Evaluation}

We evaluate document paraphrases on three aspects: Diversity, Semantic Relevancy and Coherence.

\textbf{Diversity:}
Previous works use \textbf{self-BLEU}, which calculate BLEU score between original text and generated paraphrase, to measure the diversity of paraphrase. However, we find that BLEU score may not be suitable for document-level paraphrase generation task, as it only measures the diversity of N-gram phrase and ignores the inter-sentence diversity. TER \citep{zaidan-callison-burch-2010-predicting} \footnote{The tool of TER is available at \url{https://github.com/jhclark/multeval}.} and WER \footnote{The tool of WER is available at \url{https://github.com/belambert/asr-evaluation}.} are used to evaluate machine translation and automatic speech recognition based on edit distance. Previous works also employ self-TER and self-WER to evaluate the diversity of paraphrase \citep{gupta2017deep,goyal-durrett-2020-neural}. So we add \textbf{self-TER} and \textbf{self-WER} to evaluate the document-level diversity.

\textbf{Semantic Relevancy:}
In addition to diversity, paraphrase also requires to preserve the semantic of the original input. We leverage \textbf{BERTScore} \citep{Zhang2020BERTScoreET} \footnote{The tool of BERTScore is available at \url{https://github.com/Tiiiger/bert_score}.} to evaluate the semantic similarity  between output and original document.

\textbf{Coherence:}
Unlike sentence-level paraphrase, document-level paraphrase needs to maintain the inter-sentence coherence. We propose \textbf{COH} and \textbf{COH-p} based on ALBERT \citep{Lan2020ALBERT:}\footnote{We use the huggingface Transformers \citep{wolf-etal-2020-Transformers} for ALBERT and GPT2-Large.\label{hug}} to measure the inter-sentence coherence. For a generated document paraphrase $D_g = \{ S^1_g, S^2_g, \cdots, S^{N^{'}}_g \}$ where $S^i_g$ is the $i$-th sentence, we can calculate COH and COH-p as follow:

\begin{equation}
\begin{aligned}
    \operatorname{COH} &= \mathbb{E}\left[\mathbbmss{1} \{ P_{SOP}(S_g^i, S_g^{i + 1}) \ge 0.5 \}\right] \\
    \operatorname{COH-p} &= \mathbb{E} \left[ P_{SOP}(S_g^i, S_g^{i + 1}) \right]
\end{aligned}
\end{equation}

In addition, we report perplexity for all outputs. For fairness, we employ GPT2-Large\textsuperscript{\ref{hug}} without any fine-tuning to compute the PPL score.

\subsection{Baseline}

Because there is no existing document-level paraphrasing model, we mainly adapt sentence-level paraphrasing models by paraphrasing each sentence in a document individually for comparison. The sentence-level paraphrasing models include residual LSTM \citep{prakash2016neural}, SOW-REAP\citep{goyal-durrett-2020-neural}\footnote{The code and model is available at \url{https://github.com/tagoyal/sow-reap-paraphrasing}.}, pointer generator\citep{see-etal-2017-get} and Transformer. To enhance the inter-sentence diversity, we also introduce shuffle operation to typical baseline model, which random shuffles all sentences and chooses a result with $\operatorname{COH} \ge 0.5$. We set a maximum shuffle times to avoid dead cycle. We do shuffle operation before and after Transformer-based model respectively, and use them as another two baselines. 

In addition, we leverage the pseudo document-level paraphrase dataset to directly train a document-level Transformer model (Transformer-doc) as the document-level paraphrasing baseline. For fairness, we also list the results of Transformer-doc with diversity coefficient.

The default decoding algorithm for the models is beam search. Our model and baseline models can be further integrated with top-k decoding \citep{fan-etal-2018-hierarchical} to improve the diversity.  

\subsection{Training Details}

For graph construction, we set $\epsilon = 0.5$. For diversity coefficient, we focus on the diversity of the first two grams and set $\lambda_1 = 2$, $\lambda_2 = 1$. Other hyper-parameters of our model are the same as Transformer. We selected the best hyper-parameter configuration using the highest COH score on the validation data.

\section{Results}

\subsection{Automatic Evaluation}

Table \ref{result} shows the results of automatic evaluation. Compared with other sentence-level paraphrasing models, our model gets the highest BERTScore, self-TER and self-WER. This means that our model can generate document paraphrase which is quite different from the original document while still well preserving the semantics. Although Transformer-doc gets a high BERTScore, its self-TER and self-WER scores are much lower, which means that the result of Transformer-doc is too similar with the original document. Although we integrate top-k decoding and diversity coefficient with Transformer-doc to increase diversity, this problem can not be solved well.

\begin{table*}
\centering
\small
\scalebox{0.95}{
\begin{tabular}{l|cccccccc}
\toprule[1.2pt]

\textbf{Model}       &\textbf{BERTScore}$\uparrow$&\textbf{COH}$\uparrow$&\textbf{COH-p}$\uparrow$&\textbf{self-TER}$\uparrow$&\textbf{self-WER}$\uparrow$&\textbf{self-BLEU}$\downarrow$& \textbf{PPL} \cr
\hline
Source               &   -            & 87.39          & 77.99          &   -            &  -             &   -   & 34.54 \cr
\hline
Residual LSTM        & 45.47          & 76.18          & 64.05          & 51.03          & 54.75          & 32.21 & 60.52 \cr
Pointer Generator    & 57.02          & 78.85          & 66.80          & 47.63          & 51.18          & 37.07 & 45.61 \cr
Transformer-sent     & 62.99          & \textbf{80.34} & \textbf{69.11} & 44.70          & 48.89          & 37.47 & 43.30 \cr
\quad+ top-k (k=5) & 49.94          & 71.31          & 68.75          & 57.91          & 63.20          & 24.15 & 61.85 \cr
\quad+ shuffle before& 59.83          & 56.31          & 51.19          & 54.51          & 55.32          & 36.95 & 43.55 \cr
\quad+ shuffle after & 60.41          & 59.13          & 54.02          & 55.37          & 56.07          & 36.44 & 43.41 \cr
SOW-REAP             & 64.40          & 67.58          & 61.53          & 16.68          & 31.35          & 78.26 & 76.98 \cr
Transformer-doc*     & 91.46          & 86.93          & 76.87          &  \textcolor{red}{9.36}          & \textcolor{red}{10.32}          & 83.30 & 39.84 \cr
\quad+ top-k (k=5) & 81.14         & 82.20          & 77.89          &  \textcolor{red}{21.15}         & \textcolor{red}{23.15}          & 66.17 & 51.88 \cr
\quad+shuffle before  &79.30          &\textcolor{red}{58.74}           &\textcolor{red}{55.35}           &53.59           & 58.73          &78.91 &44.19   \cr
\quad+shuffle after   &83.57          &\textcolor{red}{61.01}           &\textcolor{red}{58.45}           &51.39           & 54.87          &79.80 &45.51   \cr
\quad+div coef        &76.75          &81.10           &75.38           &\textcolor{red}{25.41}           & \textcolor{red}{28.75}          &63.58 &46.88   \cr
\hline
CoRPG(beam)          & \textbf{70.52} & 79.21          & 68.29          & 60.00          & 64.97          & 60.83 & 49.80 \cr
CoRPG(top-k, k=5)         & 59.69          & 74.19          & 63.72          & \textbf{68.92} & \textbf{74.77} & 48.62 & 67.47 \cr

\bottomrule[1.2pt]
    \end{tabular}}
\caption{\label{result} Results of automatic evaluation. The evaluation metrics include diversity, semantic relevancy and inter-sentence coherence. * indicates that the outputs of the Transformer-doc-based models are either lacking of diversity or getting lower coherence scores, which can not be seen as valid paraphrases. We mark the sick scores in red.}
\end{table*}

\begin{table*}[htb]
\centering
\small
\scalebox{0.95}{
\begin{tabular}{l|cccccccc}
\toprule[1.2pt]

\textbf{Model}       &\textbf{BERTScore}&\textbf{COH}  & \textbf{COH-p} &\textbf{self-TER}&\textbf{self-WER}&\textbf{self-BLEU}& \textbf{PPL} \cr
\hline
CoRPG                & 70.52 & 79.21 & 68.29 & 60.00 & 64.97 & 60.83 & 49.80 \cr
\hline
w/o div coef         & 78.37 & 81.73 & 69.80 & 52.85 & 56.53 & 75.75 & 40.78 \cr
w/o graph GRU        & 67.80 & 57.28 & 53.48 & 66.04 & 72.86 & 61.50 & 42.02 \cr
sent position        & 79.33 & 85.13 & 74.51 & 22.46 & 23.67 & 64.06 & 46.20 \cr
GAT                  & 64.66 & 64.90 & 59.77 & 62.87 & 69.80 & 57.94 & 40.25 \cr  

\bottomrule[1.2pt]
\end{tabular}}
\caption{\label{ablation} Results of ablation study. sent position is the model that we remove graph GRU and add positional embedding for each sentence. GAT is the model that we replace graph GRU with GAT.}
\end{table*}

In terms of inter-sentence coherence, although our model changes the order of sentences, it still gets high COH and COH-p scores, which are only a little bit lower than Transformer-sent, but much higher than other models. However, Transformer-sent with shuffle operation, which can also change sentence's order, gets low coherence and diversity scores. This means that our model can indeed improve the diversity across sentences without affecting the coherence. The ALBERT for COH calculation is fine-tuned on the training data. Therefore, Transformer without introducing inter-sentence diversity tends to get high coherence score, as its outputs are more consistent with the training data. 

Previous sentence-level paraphrasing models such as residual LSTM and SOW-REAP only focus on the diversity within a sentence. These models can generate diverse sentences but lack of inter-sentence diversity. 

\subsection{Ablation Study}

We perform ablation study to investigate the influence of different modules in our CoRPG model. We remove graph GRU and diversity coefficient respectively to explore their effect. In order to explore the effectiveness of graph GRU further, we also add two experiments. One of the experiments is that we remove graph GRU and add positional embedding for each sentence. In another experiment, we replace graph GRU with GAT. All models in ablation study employ beam search to generate paraphrase. Table \ref{ablation} shows the results of ablation study.

We can see that each module in our model does contribute to the overall performance. Diversity coefficient can increase the diversity in some degree. The model without graph GRU gets very low coherence scores. Moreover, neither sentence positional embedding nor GAT can replace graph GRU. Sentence positional embedding will reduce diversity, and GAT can not catch the coherence relationship well although it achieves excellent performance in other tasks.

In Appx.\ref{appendix:eplison}, we make a future exploration about the influence of choosing different $\epsilon$ when constructing the coherence relationship graph.  

\subsection{Human Evaluation}

We perform human evaluation on the outputs of our CoRPG model and four strong baselines in three aspects: diversity, relevancy and coherence. All ratings were obtained using a five point Likert scale. We randomly sample 100 instances from the model's outputs. We employ 6 graduate students to rate each instance, and we ensure every instance is rated by at least three judges. We also calculate kappa coefficient to measure the consistency for each judge's evaluation. More details about human evaluation are shown in Appx.\ref{appendix:human}. The results are shown in Table \ref{human}.

\begin{table}[htb]
\centering
\small
\scalebox{0.96}{
\begin{tabular}{l|ccc}
\toprule[1.2pt]

\textbf{Model}       &\textbf{Relevancy}&\textbf{Diversity}  & \textbf{Coherence} \cr
\hline
Transformer-sent     & 3.72  & 3.53  & 3.75  \cr
\quad +shuffle before& 2.86  & 3.80  & 2.83  \cr
\quad +shuffle after & 3.16  & 3.68  & 3.15  \cr
SOW-REAP             & 2.60  & 3.65  & 3.43  \cr
CoRPG                & \textbf{3.96}  & \textbf{4.12}  & \textbf{3.86}  \cr
\hline
Cohen's Kappa        & 0.581 & 0.669 & 0.625 \cr
\bottomrule[1.2pt]
\end{tabular}}
\caption{\label{human} Results of human evaluation.}
\end{table}

From the table, we can see that the outputs of our model get high score on diversity and relevancy, which means that our model can generate document paraphrases with more diversity while still preserving the semantics of original document. In addition, our model also gets the highest score on coherence even if our model may change the sentence's order in a document.
Sentence-level paraphrasing model such as Transformer-sent and SOW-SEAP can only focus on the intra-sentence diversity, but ignore the inter-sentence diversity. Simply improving the inter-sentence diversity by shuffling leads to the decrease of coherence. Because there are many sentences in a document, it can hardly find the result with good diversity and coherence through shuffle. Although the BERTScores of the Transformer-sent with ``shuffle before'' and ``shuffle after'' are high, the relevancy scores of these two models are low. This is because low coherence may lead human to feel low relevance.
Cohen's kappa of human evaluation is high enough, so we think the human evaluation is credible.

\subsection{Case Study}

We perform case studies for better understanding the model performance. Table \ref{case} shows an example of document paraphrase. Obviously, document paraphrase generated by sentence-level paraphrasing model can only rewrite individual sentences. Although it can increase lexical diversity, the sentence-level paraphrase model ignores the inter-sentence coherence and diversity. On the contrary, our CoRPG model can not only rewrite words in each sentence but also reorder sentences while still preserving its original semantic information and having good coherence.

\begin{table}
    \centering
    \scalebox{0.82}{
    \begin{tabular}{|p{9cm}|}
    \hline
    \textbf{Original:} \cr
    \textcolor{red}{$_1$}Another aspect of the pivot involves moving away from the middle east. 
    \textcolor{blue}{$_2$}But no amount of fancy footwork, whether pivoting or pirouetting, can diminish that region ’s importance. 
    \textcolor{olive}{$_3$}The middle east will remain a central pillar of world energy for decades to come, whether it ultimately can export more energy than instability is the key question.  
    \textcolor{orange}{$_4$}Unlike east asia, the middle east remains a region in turmoil, the complexity of which defies analytical consensus. 
    \textcolor{purple}{$_5$}Do the region ’s crises stem from the lack of peace with israel? \cr
    \hline
    \textbf{Transformer-sent:} \cr
    \textcolor{red}{$_1$}Another aspect of the pivot is \textcolor{blue}{to} move from the \textcolor{blue}{middle of the east}. 
    \textcolor{blue}{$_2$}But no fancy \textcolor{blue}{work}, whether \textcolor{blue}{pivoted or pivoted}, can diminish the importance of this region. 
    \textcolor{olive}{$_3$}The middle east will remain a central pillar of world energy for decades. 
    \textcolor{orange}{$_4$}Unlike east asia, the middle east is \textcolor{blue}{still} a region of turmoil, whose complexity \textcolor{blue}{is destroying} analytical consensus.
    \textcolor{purple}{$_5$}Is there a lack of peace with the \textcolor{blue}{island} of the region? \cr
    \hline
    \textbf{CoRPG:} \cr
    \textcolor{red}{$_1$}Another aspect of the pivot involves moving away from the middle east. 
    \textcolor{orange}{$_4$}Unlike east asia, the middle east remains a region in \textcolor{blue}{tumult}, complexity, defies analytical consensus. 
    \textcolor{olive}{$_3$}\textcolor{blue}{For a decade or so}, the middle east will remain the central pillar of world center \textcolor{blue}{for whom} it ultimately can export more \textcolor{blue}{of its} energy than instability is a key \textcolor{blue}{issue}. 
    \textcolor{blue}{$_2$}But no amount of fancy footwork, \textcolor{blue}{as in} pivoting and pirouetting may \textcolor{blue}{weaken} that region’s \textcolor{blue}{significance}. 
    \textcolor{purple}{$_5$}Do the region’s crises stem from \textcolor{blue}{its} lack of a peace \textcolor{blue}{deal} with israel? \cr
    \hline
    \end{tabular}}
\caption{An example for case study. The number before each sentence in the generated paraphrase indicates the corresponding original sentence from the input document. The word in color means that it does not appear in the original sentence.}
\label{case}
\end{table}

\section{Conclusion}

In this paper, we explore a challenging document-level paraphrase generation task and propose a novel model called CoRPG to generate document paraphrases with good relevancy, coherence and inter-sentence diversity.  Both automatic and human evaluation show the efficacy of our model. In the future, we will try to incorporate  the operations of sentence splitting and merging, which is not well addressed by our model, to further improve the quality of document paraphrase.

\section*{Acknowledgments}

This work was supported by National Natural Science Foundation of China (61772036), Beijing
Academy of Artificial Intelligence (BAAI) and Key Laboratory of Science, Technology and Standard
in Press Industry (Key Laboratory of Intelligent Press Media Technology). We appreciate the anonymous reviewers for their helpful comments. Xiaojun Wan is the corresponding author.

\bibliography{anthology,custom}
\bibliographystyle{acl_natbib}

\appendix
\clearpage

\section{Explanation of Diversity Coefficient}
\label{appendix::div-coef}
\begin{figure}[htb]
\centering
\includegraphics[scale=1.1]{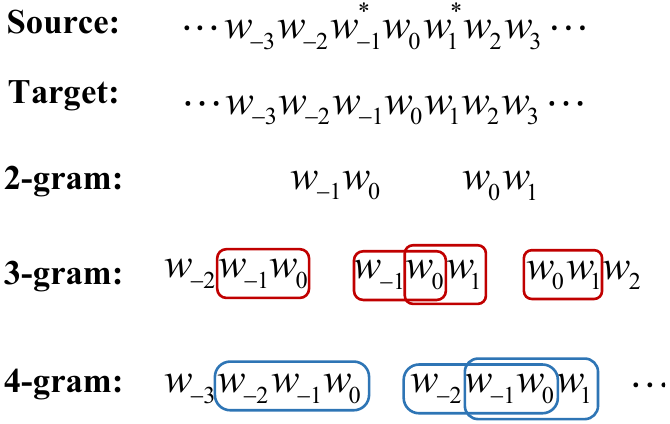} 
\caption{ A example of diversity coefficient. The red rectangle indicates the 2-gram phrase of $w_0$ contained in the 3-gram phrase of $w_0$, and the blue rectangle indicates the 3-gram phrase of $w_0$ contained in the 4-gram phrase of $w_0$.}
\label{div-coef}
\end{figure}

The motivation of the diversity coefficient is that we want the model to pay more attention to the N-gram phrase diversity. Concretely, we penalize a word if all N-gram phrases containing this word in the target sentence do not appear in the source sentence. As is shown in the Figure \ref{div-coef}, for the word $w_0$, the 2-gram phrases $w_{-1}w_0$ and $w_0w_1$ do not appear in the source sentence. So, we penalize the $w_0$ with 2-gram diversity coefficient $\lambda_2$.

However, there may exist repeated penalties. For example, in Figure \ref{div-coef}, all 3-gram phrases of $w_0$ contain some 2-gram phrases of $w_0$, and so forth. If we penalize the 2-gram phrase, other N-gram (N$>$2) phrases also satisfied the conditions of being penalized. To avoid this problem, we only penalize each word once at most. In this example, we only penalize the 2-gram situation.


\section{Dataset}
\label{appendix:dataset}
\textbf{ParaNMT} is a sentence-level paraphrase dataset which is created by back-translation between two languages. 
This dataset has been widely used in many previous works \citep{goyal-durrett-2020-neural}.
ParaNMT includes more than 50M sentence paraphrase pairs and the similarity score between original sentence and paraphrase sentence.
We leverage this dataset to train our sentence-level paraphrase model because it covers a wide range of domains. In order to balance the diversity and semantic relevance, we choose the paraphrase pairs with similarity score between 0.7 and 0.8 and self-BLEU less than 10. We also discard all sentences shorter than 10 words.

\textbf{News Commentary} is a monolingual translation dataset that includes document-level news corpus. In order to reduce the document length and increase the amount of training data, we split the full news article into short documents with five sentences. We employ off-the-shelf sentence-level paraphrase model to generate pseudo document-level paraphrase as source and leverage the original document as target. We will publish this pseudo document-level paraphrase dataset later.

Table \ref{statistic} provides statistics of these two datasets. 

\begin{table}[htb]
\centering
\small
\scalebox{0.98}{
\begin{tabular}{c|ccc}
\toprule[1.2pt]

\textbf{Dataset} & \textbf{Train Set} &\textbf{Valid Set} & \textbf{Test Set}\cr
\hline
ParaNMT         & 988,785 & 3,000 & - \cr
News Commentary & 96,889  & 3,000 & 3,000\cr
\bottomrule[1.2pt]
\end{tabular}}
\caption{Statistic for datasets: the sizes of train, valid and test sets.}
\label{statistic}
\end{table}

\section{The details of BERTScore}
\label{appendix:metrics}

BERTScore leverages Roberta to calculate the similarity between two sentences. We use default parameters provided by \citep{Zhang2020BERTScoreET}. We choose the F1 of BERTScore to evaluate our model. The higher the semantic similarity, the higher the value of BERTScore. Because the difference of BERTScores for different outputs are small, we employ ``rescale with baseline'', provided by the author, to rescale the score. This may reduce the value of BERTScore (without changing the ranking), but can make the score more intuitive.

\section{Ablation Study on $\epsilon$}
\label{appendix:eplison}

\begin{figure}[htb]
\centering
\includegraphics[scale=0.55]{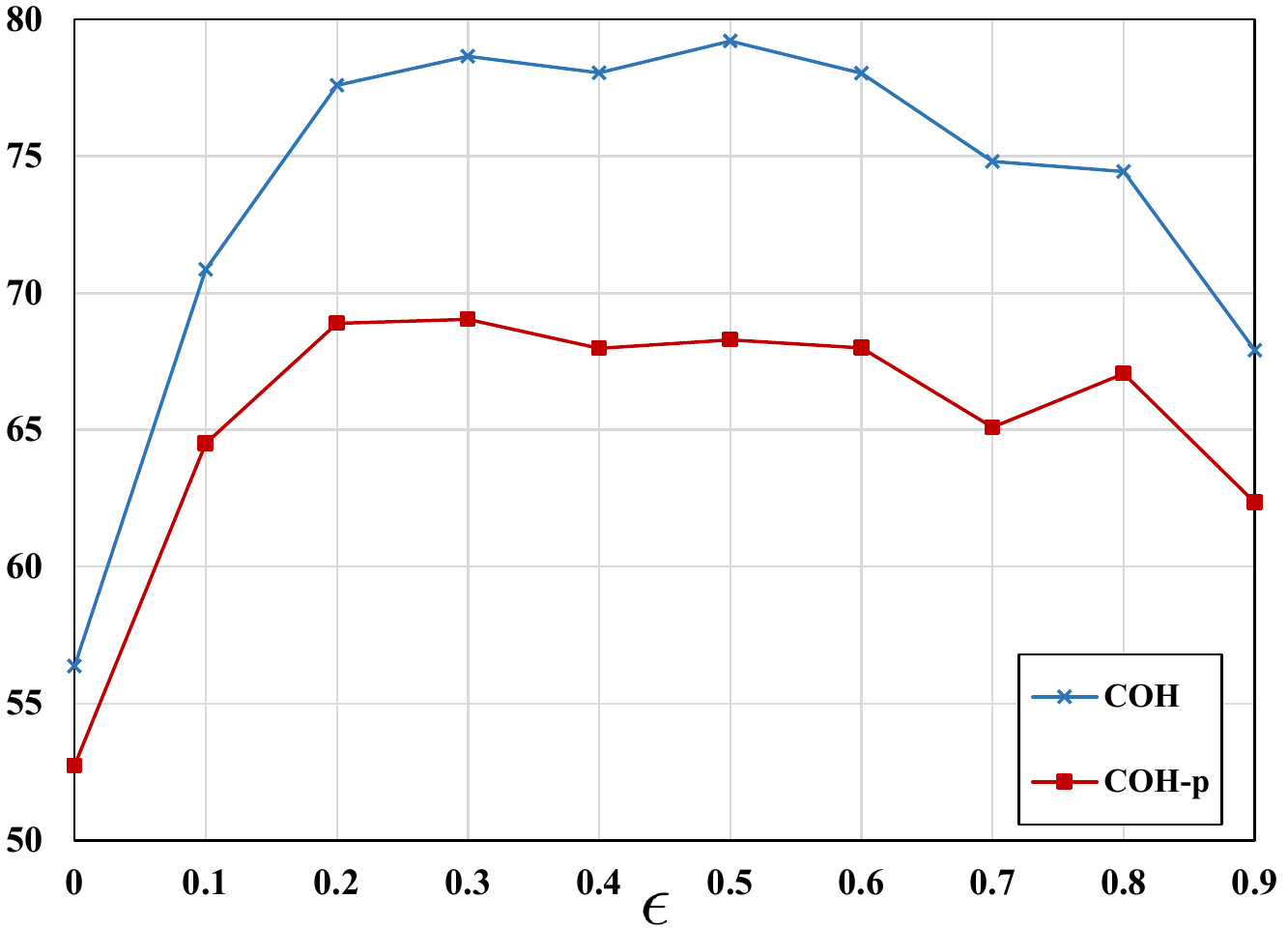}
\caption{ The curves of COH and COH-p for different $\epsilon$.}
\label{coherence}
\end{figure}

 We explore the influence of choosing different $\epsilon$ when constructing the coherence relationship graph. Figure \ref{coherence} shows the results.

As shown in the figure, with the increase of $\epsilon$, the coherence scores become higher. However, a very large $\epsilon$ may lead to the decrease of COH and COH-p. Because a very high $\epsilon$ may cause many isolated nodes, which means that $\mathbb{G}(i, \cdot) = 0$ and $\mathbb{G}(\cdot, i) = 0$. Too many isolated nodes will lead to the loss of coherence relationship.

\section{Human Evaluation}
\label{appendix:human}

We perform human evaluation of model's outputs with respect to three parts: diversity, relevancy and coherence. 
\begin{itemize}
    \item For diversity, we require judges to evaluate how much difference there is between paraphrase and original document.
    
    \item For relevancy, judges need to judge whether the semantics of the generated paraphrase are similar to the original document.

    \item For coherence, judges need to evaluate two aspects. One is the fluency of each sentence in the paraphrase document. Another is the inter-sentence coherence and consistency.

\end{itemize}

We put all outputs of different models together and let judges rate them in diversity and relevancy by comparing with the original documents. For coherence, because the original document may affect the judgment of judges, we require judges to rate a single text each time (without seeing and comparing with the original document). The sampled instances used in coherence evaluation are the same as those used in diversity and relevancy evaluation.

\end{document}